\documentclass[sigconf]{acmart}
\AtBeginDocument{%
  \providecommand\BibTeX{{%
    \normalfont B\kern-0.5em{\scshape i\kern-0.25em b}\kern-0.8em\TeX}}}

\usepackage{comment}

\copyrightyear{2021}
\acmYear{2021}
\setcopyright{acmcopyright}
\acmConference[FIRE 2021]{Forum for Information Retrieval Evaluation}{December 13--17, 2021}{Virtual Conference, India}
\acmBooktitle{Forum for Information Retrieval Evaluation (FIRE 2021), December 13--17, 2021,  Virtual Conference  , India}
\acmPrice{15.00}
\acmDOI{10.1145/3503162.3505240}
\acmISBN{ 978-1-4503-9596-0/21/12}

\addtocounter{page}{36}
\begin{document}

%%
%% The "title" command has an optional parameter,
%% allowing the author to define a "short title" to be used in page headers.
\title{UrduFake@FIRE2021: Shared Track on \\ Fake News Identification in Urdu}

%%
%% The "author" command and its associated commands are used to define
%% the authors and their affiliations.
%% Of note is the shared affiliation of the first two authors, and the
%% "authornote" and "authornotemark" commands
%% used to denote shared contribution to the research.
\author{Maaz Amjad}
%%\authornote{Both authors contributed equally to this research.}
%%\orcid{1234-5678-9012}
\affiliation{%
  \institution{ Instituto Polit\'ecnico Nacional, Center for Computing Research (CIC)}
  \city{Mexico City}
  \country{Mexico}
} \email{maazamjad@phystech.edu}

\author{Sabur Butt}
\affiliation{%
  \institution{ Instituto Polit\'ecnico Nacional, Center for Computing Research (CIC)}
  \city{Mexico City}
  \country{Mexico}}
\email{sabur@nlp.cic.ipn.mx}

\author{Hamza Imam Amjad}
\affiliation{%
  \institution{Moscow Institute of Physics and Technology}
  \city{Moscow}
  \country{Russia}}
\email{hamzaimamamjad@phystech.edu}

\author{Grigori Sidorov}
\affiliation{%
  \institution{ Instituto Polit\'ecnico Nacional, Center for Computing Research (CIC)}
  \city{Mexico City}
  \country{Mexico}}
\email{sidorov@cic.ipn.mx}

\author{Alisa Zhila}
\affiliation{%
  \institution{Ronin Institute for Independent Scholarship}
  \city{}
  \country{United States}}
\email{alisa.zhila@ronininstitute.org}

\author{Alexander Gelbukh}
\affiliation{%
  \institution{ Instituto Polit\'ecnico Nacional, Center for Computing Research (CIC)}
  \city{Mexico City}
  \country{Mexico}}
\email{gelbukh@gelbukh.com}

%%
%% By default, the full list of authors will be used in the page
%% headers. Often, this list is too long, and will overlap
%% other information printed in the page headers. This command allows
%% the author to define a more concise list
%% of authors' names for this purpose.
\renewcommand{\shortauthors}{Amjad, Sidorov et al.}

%%
%% The abstract is a short summary of the work to be presented in the
%% article.
\begin{abstract}
This study reports the second shared task named as UrduFake@Fire2021 on identifying fake news detection in Urdu language. This is a binary classification problem in which the task is to classify a given news article into two classes: (i) real news, or (ii) fake news. In this shared task, 34 teams from 7 different countries (China, Egypt, Israel, India, Mexico, Pakistan, and UAE) registered to participate in the shared task, 18 teams submitted their experimental results and 11 teams submitted their technical reports. The proposed systems were based on various count-based features and used different classifiers as well as neural network architectures. The stochastic gradient descent (SGD) algorithm outperformed other classifiers and achieved 0.679 F-score. 
\end{abstract}

%%
%% The code below is generated by the tool at http://dl.acm.org/ccs.cfm.
%% Please copy and paste the code instead of the example below.
%%

%\ccsdesc{Computer systems organization~Robotics}
%\ccsdesc[100]{Networks~Network reliability}

%%
%% Keywords. The author(s) should pick words that accurately describe
%% the work being presented. Separate the keywords with commas.

%% A "teaser" image appears between the author and affiliation
%% information and the body of the document, and typically spans the
%% page.
% \begin{teaserfigure}
%   \includegraphics[width=\textwidth]{sampleteaser}
%   \caption{Seattle Mariners at Spring Training, 2010.}
%   \Description{Enjoying the baseball game from the third-base
%   seats. Ichiro Suzuki preparing to bat.}
%   \label{fig:teaser}
% \end{teaserfigure}

%%
%% This command processes the author and affiliation and title
%% information and builds the first part of the formatted document.
\maketitle

\section{Introduction}

Fake news is used as a tool to sway the opinion of people that has a direct impact on society, economy, politics, health and various other domains of life. The COVID-19 pandemic brought a sea of fake news generated through social media, for example, influencing people from getting vaccinated~\cite{van2020inoculating}. Similarly, in the political domain, we saw Indian media reporting fabricated news of the outbreak of civil war in Karachi, Pakistan~\footnote{https://www.bbc.com/news/world-asia-54649302}. The incidents became frequent and need timely action to prevent social unrest. As much as fake news affects the rest of the world, the South Asian people had their fair share of fake news exposure. Hence, there exists a vacuum for fake news study in low resource languages, for example, Urdu, which is spoken by more than 230 million people~\footnote{\url{https://www.statista.com/statistics/266808/the-most-spoken-languages-worldwide/}} in South Asia and worldwide. 

Automatic detection of fake news in English has many textual approaches to classify fake news in various forms. NLP experts used supervised machine learning algorithms (Random Forest (RF) \cite{amjad2020data}, Support Vector Machines (SVM) \cite{amjad2020data}, Decision Trees, etc.)~\cite{ashraf2021cic, wu2015false, amjad2020data} and deep learning algorithms (Long short-term memory (LSTM), Recurrent Neural networks (RNN), Gated Recurrent Units (GRU), etc.)~\cite{socher2012semantic, roy2018deep} with the combination of various embeddings, linguistic and user-level features. Though many of these approaches are successful in the English language, Urdu has a different morphological and syntactic structure and lacks resources for many deep learning approaches. Hence, it is important to continue developing resources and methods to reach real-life deployable solutions.

This paper gives a summary of the UrduFake track at FIRE 2021, which is in the continuation of its 2020 version. The dataset, methodologies and baselines changed with the expansion of the dataset between these two tasks. A detailed explanation of the UrduFake track at FIRE 2021~\cite{amjad2021urdufake} and 2020~\cite{6} is available for the research community. The goal is this study is to generate more resources for fake news classification in the Urdu language. We encouraged participation from all over the world and the submitted methodologies are compiled and explained in Section~\ref{overview}. We summarized the challenges, differences and future direction for combating fake news distribution on digital media platforms.

\section{Task Description}
The binary classification task remained unchanged since the last edition~\cite{6, amjad2020urdufake}. The teams were required to assign a label (real or fake news) for given news articles written in Urdu Nastal\'{i}q script. The used definition of fake news is presented in the overview and dataset articles~\cite{6, 11}. 

Fake News Detection: For an unannotated news article, denoted as  $\alpha$, where $\alpha$  $\in$ $N$ ($N$ represents the total articles), an automatic fake news detection algorithm assigns a score $S(\alpha)$ $\in$ [0, 1]. Given that $S(\widehat{\alpha})$ > $S(\alpha)$, we predict that $\widehat{\alpha}$ has higher probability to be a fake news article. A threshold $\gamma$ can be defined, such that the prediction function $F$ : $N$ $\rightarrow$ 
[not fake, fake] is:
\begin{align*}
F(N) &=  \begin{cases}
fake, \,\,\,\,\,\,if \,\,\,\,\ S(\alpha) \in \gamma), \\
not fake, \,\,\,\,\,\,\,\,\  \textrm{otherwise}.
\end{cases}
\end{align*}

\section{Data Collection and Annotation }
The dataset used in this shared task  contained five types of news articles: (i) Business, (ii) Health, (iii) Showbiz (entertainment), (iv) Sports, and (v) Technology. To assemble the dataset, numerous news articles were crawled from well-known traditional media news streams (national and international) using Python library Newspaper\footnote{https://newspaper.readthedocs.io/en/latest/}. The Newspaper library automatically removes unimportant information, such as author’s name, date, advertisements, location of the publisher, HTML tags and images etc. The dataset is publically available for research purposes\footnote{ https://github.com/MaazAmjad/Urdu-Fake-news-location FIRE2021.git}.

\begin{itemize}
\item \textit{\textbf{Real News Collection for Training and Testing Sets}}:\\ 
All the real news were retrieved from news agencies \cite{11} and were manually annotated using a set of guidelines. For example, if the information mentioned in a news article can be verified from other news sources, then the news article can be annotated as a real news article. The detailed instruction of how the real news were annotated are reported in our earlier research \cite{11}. \\

\item \textit{\textbf{Fake  News Collection by Professional  Crowdsourcing}}:\\ 
The fake news used in this shared task were written by professional journalists. The real news were provided to the professional journalists and they were asked to write corresponding fake news articles. Professional crowdsourcing was used because collecting corresponding fake news is a challenging task. The detailed instruction of how the fake news were collected are reported in our earlier research \cite{11}.  
\end{itemize}

Table {\ref{tab:freqq}} describes the distribution of the news articles in the dataset. 

\begin{table}[!htbp]
  \caption{Distribution of the news articles.}
  \label{tab:freqq}
    \resizebox{5cm}{1cm}{
  \begin{tabular}{cccl}
    \toprule
    &Real News  & Fake News  & Total\\
    \midrule
    \textbf{Train} &  750  & 550  & 1,300\\
    \textbf{Test}  &  200  & 100  & 300\\
  \bottomrule
 \textbf{ Total}   & 950   & 650  & 1,600 \\
  \bottomrule
\end{tabular}
}
\end{table}

\section{Evaluation Metrics}
The shared task presented a challenge to classify a news article as fake or real. The participants were provided with the train and development dataset at the beginning of the competition. The test dataset was later revealed for the teams to evaluate their performances. All the participants were only allowed three attempts to submit their best performing models. 

The test labels provided by the participants were compared with the ground truth labels. The standard evaluation metrics for fake news: Recall (R), Precision (P), Accuracy, and two F1-scores (F1-score for each class and F1-macro) were used. The F1-score has multiple variants like F1-macro, weighted F1,  or F1-micro. To calculate the label of the “real” class F1\textsubscript{real} was used and the label of the “fake” class out of all news was calculated through F1\textsubscript{fake}. 

The macro-averaged F1-macro, which is the average of F1\textsubscript{real} and F1\textsubscript{fake} was used to tackle the imbalance of the dataset. F1-macro penalizes when a system does not perform well for the minority classes and does not use weights for the aggregation. It should be noted, that we only report F1-macro. This is because when the weighted F1 of both classes is summed up, it gives more weight to the majority class.

\section{Baseline Systems}
A baseline was proposed to serve as reference points so that the proposed systems can be evaluated and ranked them accordingly. We used bag of words (BoW) model and $n$-gram (char, word) features and trained different machine learning classifiers. Overall, Decision Tree classifier provided the best results using bi-grams of the combination of char-word-function words with tf-idf. 

\section{Overview of the Submitted Approaches}~\label{overview}
The submitted approaches spanned from machine learning methods to deep learning and transformer methods. 34 teams from 6 different countries (India, Pakistan, China, Egypt, Germany, and the UK) registered for participation, 18 teams submitted their experimental results and from which 11 teams submitted their technical reports. Registered participants were from 6 different countries (India, Pakistan, China, Egypt, Germany, and the UK). The results of the proposed approaches did not prove to be better than the baseline methods, except for two participant teams. Table~\ref{tab:freq} shows the approaches used by the teams and table~\ref{tab:overlapping2} presents the best run scores achieved using those methods.

%This section gives a brief overview of the systems submitted to this competition. 42 teams registered for participation, from which 9 teams submitted their runs. Registered participants were from 6 different countries (India, Pakistan, China, Egypt, Germany, and the UK). Participation of different teams from multiple countries confirms the importance of this task. The team members came from various types of organizations: universities, research centres, and industry. Table {\ref{tab:freq}} describes the submitted approaches.

\section{Conclusion} 
This shared task aims to attract researcher to address the fake news detection in Urdu language. A first news dataset in Urdu language has been proposed for fake news detection task which contains 1,600 news articles. All the real news articles were crawled from national and international reliable sources and the information in these articles was manually verified. On the other hand, the professional journalists were hired for the corresponding fake news generation. In this shared task, the Stochastic Gradient Descent algorithm outperformed all the classifiers and obtained the highest score in identifying fake news in Urdu. We aim to increase the size of the news dataset in Urdu and use deep learning learning techniques to identify fake news in Urdu language.

\begin{table}[]
	\caption{Participants’ best run scores. }
	  \resizebox{\columnwidth}{!}{%
	\label{tab:overlapping2}
	  \centering
\begin{tabular}{llllllllllll}
\noalign{\smallskip}\hline\noalign{\smallskip}
 & \multicolumn{4}{c}{\textbf{Fake Class}}                              & \multicolumn{4}{c}{\textbf{Real Class}}                             & \multicolumn{2}{l}{}          &          \\
 \textbf{Team Names} & Prec              & Recall            & \multicolumn{2}{l}{F1\_Fake} & Prec             & Recall            & \multicolumn{2}{l}{F1\_Real} & \multicolumn{2}{l}{F1\_Macro} & Accuracy \\
\noalign{\smallskip}\hline\noalign{\smallskip}
\textbf{ Nayel }                                                       & 0.754             & 0.400             & \multicolumn{2}{l}{0.522}                            & 0.757            & 0.935             & \multicolumn{2}{l}{0.836}                            & \multicolumn{2}{l}{0.679}                             & 0.756    \\
 \noalign{\smallskip}\hline\noalign{\smallskip}
\textbf{ Abdullah-Khurem  }                                            & 0.592             & 0.480             & \multicolumn{2}{l}{0.530}                            & 0.762            & 0.835             & \multicolumn{2}{l}{0.797}                            & \multicolumn{2}{l}{0.663}                             & 0.716    \\
 \noalign{\smallskip}\hline\noalign{\smallskip}
 \hline
\textbf{ Baseline }                            & 0.584             & 0.450             & \multicolumn{2}{l}{0.508}                            & 0.753            & 0.840              & \multicolumn{2}{l}{0.794}                            & \multicolumn{2}{l}{0.651}                             & 0.710    \\
 \noalign{\smallskip}\hline\noalign{\smallskip}
 \hline
\textbf{ Hammad-Khurem }                                           & 0.634             & 0.330             & \multicolumn{2}{l}{0.434}                            & 0.729            & 0.905             & \multicolumn{2}{l}{0.808}                            & \multicolumn{2}{l}{0.621}                             & 0.713    \\
 \noalign{\smallskip}\hline\noalign{\smallskip}
\textbf{ Muhammad Homayoun }                                            & 0.480             & 0.490             & \multicolumn{2}{l}{0.485}                            & 0.742            & 0.735             & \multicolumn{2}{l}{0.738}                            & \multicolumn{2}{l}{0.611}                             & 0.653    \\
\noalign{\smallskip}\hline\noalign{\smallskip}
\textbf{Snehaan bhawal}                                               & 0.960             & 0.240             & \multicolumn{2}{l}{0.384}                            & 0.723            & 0.995             & \multicolumn{2}{l}{0.837}                            & \multicolumn{2}{l}{0.610}                             & 0.743    \\
\noalign{\smallskip}\hline\noalign{\smallskip}
\textbf{MUCIC}                                                      & 0.821             & 0.230             & \multicolumn{2}{l}{0.359}                            & 0.716            & 0.975             & \multicolumn{2}{l}{0.826}                            & \multicolumn{2}{l}{0.592}                             & 0.726    \\
\noalign{\smallskip}\hline\noalign{\smallskip}
 
  \textbf{SOA NLP}                                   &    0.793          &     0.230          & \multicolumn{2}{l}{0.356}                            &      0.356       &     0.715         & \multicolumn{2}{l}{0.823}                            &  \multicolumn{2}{l}{0.590}                             &  0.590   \\
  \noalign{\smallskip}\hline\noalign{\smallskip}
 
 \textbf{Dinamore \& Elyasafdi \_SVC }                                 & 0.720             & 0.180              & \multicolumn{2}{l}{0.288}                            & 0.701            & 0.965             & \multicolumn{2}{l}{0.812}                            & \multicolumn{2}{l}{0.550}                             & 0.703    \\
 \noalign{\smallskip}\hline\noalign{\smallskip}

\textbf{MUCS}        & 0.850             & 0.170             & \multicolumn{2}{l}{0.283}                            & 0.703            & 0.985             & \multicolumn{2}{l}{0.820}                            & \multicolumn{2}{l}{0.552}                             & 0.713    \\

\noalign{\smallskip}\hline\noalign{\smallskip}
\textbf{Iqra Ameer}                                                  & 0.454             & 0.100             & \multicolumn{2}{l}{0.163}                            & 0.676            & 0.940             & \multicolumn{2}{l}{0.786}                            & \multicolumn{2}{l}{0.475}                             & 0.660    \\
 \noalign{\smallskip}\hline\noalign{\smallskip}
\textbf{Sakshi kalra}                                                & 0.266             & 0.120             & \multicolumn{2}{l}{0.165}                           & 0.654            & 0.835             & \multicolumn{2}{l}{0.734}                            & \multicolumn{2}{l}{0.449}                             & 0.596    \\
\noalign{\smallskip}\hline\noalign{\smallskip}
 
\end{tabular}}
\end{table}

\begin{table}[]
  \caption{Approaches used by the participating systems.}
     \label{tab:freq}
  \resizebox{\columnwidth}{!}
  {%

  \begin{tabular}{ccccc}
    \toprule
System/Team Name & Feature Type & Feature Weighting Scheme & Classifying algorithm & NN-based\\
    \midrule
    
Nayel & tri-gram  &  TF-IDF  &  Stochastic Gradient Descent & No \\
Abdullah-Khurem &  N/A &  TF-IDF, Word2Vec, GloVe, fastText    &  textCNN   & Yes \\
Hammad-Khurem & n-gram   &  N/A   & Assemble (XG-Boost, Light GBM, Adaboost)    & No \\
Muhammad Homayoun  & word n-grams (1-4), char-gram (2-6)   &  CNN (4-channels)   & CNN & Yes \\
Snehaan Bhawal  & N/A   & embeddings   & MuRIL    & Yes \\
MUCIC & char-gram (1-3)   & TF-IDF    & Assemble (LSVM, LR, MLP, XGB, RF)    & No \\
SOA NLP & char-gram (1-3)   & TF-IDF    &  Dense neural networks   & Yes \\
Dinamore\&Elyasafdi\_SVC  & char tri-grams   &  TF-IDF   & SVC & No\\
MUCS & word uni-grams, char-grams (2-3)   & TF-IDF, fastText    & Assemble (MLP, ADB, GB, RF)    & No \\
Iqra Ameer & N/A   &  embeddings   & BERT-base    & Yes \\
Sakshi Kalra & N/A   &  embeddings   & RoBERTa-urdu-small  & Yes \\

  \bottomrule
\end{tabular}
}
\end{table}

%In this shared task, 18 teams participated and the team ($Nayel$) achieved the first position. Only two participating teams cross the proposed baseline. It is challenging to mention that whether the proposed systems can be used “in the wild” to address the task of automatic fake news detection. More research is required and a bigger dataset should be proposed to ensure the scalability of the proposed systems. 

% \begin{table*}
%   \caption{Some Typical Commands}
%   \label{tab:commands}
%   \begin{tabular}{ccl}
%     \toprule
%     Command &A Number & Comments\\
%     \midrule
%     \texttt{{\char'134}author} & 100& Author \\
%     \texttt{{\char'134}table}& 300 & For tables\\
%     \texttt{{\char'134}table*}& 400& For wider tables\\
%     \bottomrule
%   \end{tabular}
% \end{table*}

\begin{acks}
This competition was organized with the support from the Mexican Government through the grant A1-S-
47854 of the CONACYT, Mexico and grants 20211784, 20211884, and 20211178 of the
Secretaría de Investigación y Posgrado of the Instituto Politécnico Nacional, Mexico.

\end{acks}

%%
%% The next two lines define the bibliography style to be used, and
%% the bibliography file.
\bibliographystyle{ACM-Reference-Format}
\bibliography{main}

%%% -*-BibTeX-*-
%%% Do NOT edit. File created by BibTeX with style
%%% ACM-Reference-Format-Journals [18-Jan-2012].

\begin{thebibliography}{10}

%%% ====================================================================
%%% NOTE TO THE USER: you can override these defaults by providing
%%% customized versions of any of these macros before the \bibliography
%%% command.  Each of them MUST provide its own final punctuation,
%%% except for \shownote{}, \showDOI{}, and \showURL{}.  The latter two
%%% do not use final punctuation, in order to avoid confusing it with
%%% the Web address.
%%%
%%% To suppress output of a particular field, define its macro to expand
%%% to an empty string, or better, \unskip, like this:
%%%
%%% \newcommand{\showDOI}[1]{\unskip}   % LaTeX syntax
%%%
%%% \def \showDOI #1{\unskip}           % plain TeX syntax
%%%
%%% ====================================================================

\ifx \showCODEN    \undefined \def \showCODEN     #1{\unskip}     \fi
\ifx \showDOI      \undefined \def \showDOI       #1{#1}\fi
\ifx \showISBNx    \undefined \def \showISBNx     #1{\unskip}     \fi
\ifx \showISBNxiii \undefined \def \showISBNxiii  #1{\unskip}     \fi
\ifx \showISSN     \undefined \def \showISSN      #1{\unskip}     \fi
\ifx \showLCCN     \undefined \def \showLCCN      #1{\unskip}     \fi
\ifx \shownote     \undefined \def \shownote      #1{#1}          \fi
\ifx \showarticletitle \undefined \def \showarticletitle #1{#1}   \fi
\ifx \showURL      \undefined \def \showURL       {\relax}        \fi
% The following commands are used for tagged output and should be
% invisible to TeX
\providecommand\bibfield[2]{#2}
\providecommand\bibinfo[2]{#2}
\providecommand\natexlab[1]{#1}
\providecommand\showeprint[2][]{arXiv:#2}

\bibitem[\protect\citeauthoryear{Amjad, Butt, Amjad, Zhila, Sidorov, and
  Gelbukh}{Amjad et~al\mbox{.}}{2021}]%
        {amjad2021urdufake}
\bibfield{author}{\bibinfo{person}{Maaz Amjad}, \bibinfo{person}{Sabur Butt},
  \bibinfo{person}{Hamza~Imam Amjad}, \bibinfo{person}{Alisa Zhila},
  \bibinfo{person}{Grigori Sidorov}, {and} \bibinfo{person}{Alexander
  Gelbukh}.} \bibinfo{year}{2021}\natexlab{}.
\newblock \showarticletitle{Overview of the {shared Task} on {Fake News
  Detection} in {Urdu} at {FIRE} 2021}.
\newblock \bibinfo{journal}{\emph{CEUR Workshop Proceedings}}.
\newblock


\bibitem[\protect\citeauthoryear{Amjad, Sidorov, and Zhila}{Amjad
  et~al\mbox{.}}{2020a}]%
        {amjad2020data}
\bibfield{author}{\bibinfo{person}{Maaz Amjad}, \bibinfo{person}{Grigori
  Sidorov}, {and} \bibinfo{person}{Alisa Zhila}.}
  \bibinfo{year}{2020}\natexlab{a}.
\newblock \showarticletitle{Data augmentation using machine translation for
  fake news detection in the {U}rdu language}. In
  \bibinfo{booktitle}{\emph{Proceedings of The 12th Language Resources and
  Evaluation Conference}}. \bibinfo{pages}{2537--2542}.
\newblock


\bibitem[\protect\citeauthoryear{Amjad, Sidorov, Zhila, Gelbukh, and
  Rosso}{Amjad et~al\mbox{.}}{2020b}]%
        {6}
\bibfield{author}{\bibinfo{person}{Maaz Amjad}, \bibinfo{person}{Grigori
  Sidorov}, \bibinfo{person}{Alisa Zhila}, \bibinfo{person}{Alexander Gelbukh},
  {and} \bibinfo{person}{Paolo Rosso}.} \bibinfo{year}{2020}\natexlab{b}.
\newblock \showarticletitle{Overview of the {shared Task} on {Fake News
  Detection} in {Urdu} at {FIRE} 2020}.
\newblock \bibinfo{journal}{\emph{CEUR Workshop Proceedings}}
  (\bibinfo{year}{2020}).
\newblock
\newblock
\shownote{Working Notes of the Forum for Information Retrieval Evaluation (FIRE
  2020), Hyderabad, India.}


\bibitem[\protect\citeauthoryear{Amjad, Sidorov, Zhila, Gelbukh, and
  Rosso}{Amjad et~al\mbox{.}}{2020c}]%
        {amjad2020urdufake}
\bibfield{author}{\bibinfo{person}{Maaz Amjad}, \bibinfo{person}{Grigori
  Sidorov}, \bibinfo{person}{Alisa Zhila}, \bibinfo{person}{Alexander Gelbukh},
  {and} \bibinfo{person}{Paolo Rosso}.} \bibinfo{year}{2020}\natexlab{c}.
\newblock \showarticletitle{{UrduFake@FIRE2020: Shared Track} on {Fake News
  Detection} in {Urdu}}. In \bibinfo{booktitle}{\emph{Proceedings of the 12th
  Forum for Information Retrieval Evaluation (FIRE 2020), Hyderabad, India}}.
\newblock


\bibitem[\protect\citeauthoryear{Amjad, Sidorov, Zhila, G\'{o}mez-Adorno,
  Voronkov, and Gelbukh}{Amjad et~al\mbox{.}}{2020d}]%
        {11}
\bibfield{author}{\bibinfo{person}{Maaz Amjad}, \bibinfo{person}{Grigori
  Sidorov}, \bibinfo{person}{Alisa Zhila}, \bibinfo{person}{Helena
  G\'{o}mez-Adorno}, \bibinfo{person}{Ilia Voronkov}, {and}
  \bibinfo{person}{Alexander Gelbukh}.} \bibinfo{year}{2020}\natexlab{d}.
\newblock \showarticletitle{Bend the {T}ruth: A benchmark dataset for fake news
  detection in {U}rdu and its evaluation}.
\newblock \bibinfo{journal}{\emph{Journal of Intelligent \& Fuzzy Systems}}
  \bibinfo{volume}{39}, \bibinfo{number}{2} (\bibinfo{year}{2020}),
  \bibinfo{pages}{2457--2469}.
\newblock
\urldef\tempurl%
\url{https://doi.org/10.3233/JIFS-179905}
\showDOI{\tempurl}


\bibitem[\protect\citeauthoryear{Ashraf, Butt, Sidorov, and Gelbukh}{Ashraf
  et~al\mbox{.}}{2021}]%
        {ashraf2021cic}
\bibfield{author}{\bibinfo{person}{Noman Ashraf}, \bibinfo{person}{Sabur Butt},
  \bibinfo{person}{Grigori Sidorov}, {and} \bibinfo{person}{Alexander
  Gelbukh}.} \bibinfo{year}{2021}\natexlab{}.
\newblock \showarticletitle{CIC at CheckThat! 2021: Fake News detection Using
  Machine Learning And Data Augmentation}. CLEF.
\newblock


\bibitem[\protect\citeauthoryear{Roy, Basak, Ekbal, and Bhattacharyya}{Roy
  et~al\mbox{.}}{2018}]%
        {roy2018deep}
\bibfield{author}{\bibinfo{person}{Arjun Roy}, \bibinfo{person}{Kingshuk
  Basak}, \bibinfo{person}{Asif Ekbal}, {and} \bibinfo{person}{Pushpak
  Bhattacharyya}.} \bibinfo{year}{2018}\natexlab{}.
\newblock \showarticletitle{A deep ensemble framework for fake news detection
  and classification}.
\newblock \bibinfo{journal}{\emph{arXiv preprint arXiv:1811.04670}}
  (\bibinfo{year}{2018}).
\newblock


\bibitem[\protect\citeauthoryear{Socher, Huval, Manning, and Ng}{Socher
  et~al\mbox{.}}{2012}]%
        {socher2012semantic}
\bibfield{author}{\bibinfo{person}{Richard Socher}, \bibinfo{person}{Brody
  Huval}, \bibinfo{person}{Christopher~D Manning}, {and}
  \bibinfo{person}{Andrew~Y Ng}.} \bibinfo{year}{2012}\natexlab{}.
\newblock \showarticletitle{Semantic compositionality through recursive
  matrix-vector spaces}. In \bibinfo{booktitle}{\emph{Proceedings of the 2012
  joint conference on empirical methods in natural language processing and
  computational natural language learning}}. \bibinfo{pages}{1201--1211}.
\newblock


\bibitem[\protect\citeauthoryear{van Der~Linden, Roozenbeek, and Compton}{van
  Der~Linden et~al\mbox{.}}{2020}]%
        {van2020inoculating}
\bibfield{author}{\bibinfo{person}{Sander van Der~Linden}, \bibinfo{person}{Jon
  Roozenbeek}, {and} \bibinfo{person}{Josh Compton}.}
  \bibinfo{year}{2020}\natexlab{}.
\newblock \showarticletitle{Inoculating against fake news about COVID-19}.
\newblock \bibinfo{journal}{\emph{Frontiers in psychology}}
  \bibinfo{volume}{11} (\bibinfo{year}{2020}), \bibinfo{pages}{2928}.
\newblock


\bibitem[\protect\citeauthoryear{Wu, Yang, and Zhu}{Wu et~al\mbox{.}}{2015}]%
        {wu2015false}
\bibfield{author}{\bibinfo{person}{Ke Wu}, \bibinfo{person}{Song Yang}, {and}
  \bibinfo{person}{Kenny~Q Zhu}.} \bibinfo{year}{2015}\natexlab{}.
\newblock \showarticletitle{False rumors detection on sina weibo by propagation
  structures}. In \bibinfo{booktitle}{\emph{2015 IEEE 31st international
  conference on data engineering}}. IEEE, \bibinfo{pages}{651--662}.
\newblock


\end{thebibliography}

%%
%% If your work has an appendix, this is the place to put it.
\appendix

\end{document}